# 厦門大學

## 本 科 毕 业 论 文
（主修专业）

## 基于卷积神经网络的上市公司财务舞弊识别及可解释性研究

Financial Fraud Identification and Interpretability Study for Listed Companies Based on Convolutional Neural Network

姓　　　名：李骁

学　　　院：经济学院

　　　　系：统计学与数据科学系

专　　　业：数据科学与大数据技术

年　　　级：2020

校内指导教师：郭红丽　副教授

二〇二四 年 六 月

# 摘要

自从出现股份制公司以来，为了骗取投资者、债权人等利益相关者的信任，上市公司财务舞弊现象就常有发生。上市公司财务舞弊对于资本市场的危害极大，但由于舞弊手段隐蔽，审计人力成本高、周期长等原因，财务舞弊的识别并非易事。传统统计方法可解释性强，但难以捕捉多个特征之间的非线性关系；机器学习方法性能强大，但很难解释模型内部逻辑。此外，现有机器学习方法大多都依赖当年财务数据来识别当年是否舞弊，时效性较差。

本文旨在探索适用于大数据时代的财务舞弊识别方案，并解决"黑盒"模型可解释性差的问题。本文构建了基于卷积神经网络的财务舞弊识别模型，提出了处理面板数据的特征工程方案，对中国 A 股上市公司舞弊情况进行事前预测，并对模型进行局部解释。研究结果显示卷积神经网络在准确率、鲁棒性、时效性三方面均优于逻辑回归、LightGBM 模型，同时分类阈值在财务舞弊识别等高风险领域具有重要作用。本文从对象、特征、时间三维度进行可解释性研究，研究发现偿债能力、比率结构、治理结构、内部控制环境这四个二级指标对于舞弊识别有普适性作用，但环境指标只对某些涉及高污染的行业作用较大。研究还发现卷积神经网络在判定非舞弊与舞弊样本时的逻辑差异：非舞弊企业特征作用模式相似，起到主要作用的二级指标为比率结构、经营能力、社会责任、治理结构，而舞弊企业根据舞弊类型差异较大；非舞弊样本重要特征的各年份数据对于舞弊识别都有作用，而舞弊样本的重要特征往往聚集作用在某一个小范围的时间区域。本文分析了典型案例冠农股份，分析结果显示，对于冠农股份 2022 年舞弊行为的判定，贡献最大的二级指标为现金流分析、社会责任、治理结构、每股指标，与其实际舞弊手段相符。

**关键词**：财务舞弊识别，卷积神经网络，可解释人工智能，深度学习可视化

# 摘要

# Abstract

Since the emergence of joint-stock companies, financial fraud in listed companies has often occurred in order to gain credibility by cheating investors, creditors and other stakeholders. Financial fraud of listed companies is extremely harmful to the capital market, but due to hidden fraudulent tactics, high labor and time costs of auditing, the identification of financial fraud is not an easy task. Traditional statistical methods are highly interpretable, but can hardly capture the non-linear relationship between multiple features; machine learning methods are powerful, but it is difficult to explain the internal logic of the model. In addition, most of the existing machine learning methods identify whether the current year is fraudulent or not based on the current year's financial data, so the timeliness is poor.

This paper is to explore the financial fraud identification scheme applicable to the era of big data, and to solve the problem of poor interpretability of the "black box" model. In this paper, we construct a financial fraud identification model based on convolutional neural network, propose a feature engineering scheme to deal with panel data, predict the fraud situation of Chinese A-share listed companies in advance, and locally interpret the model. The results show that convolutional neural network is better than logistic regression and LightGBM model in terms of accuracy, robustness, and timeliness, meanwhile, classification threshold plays an important role in high-risk areas such as financial fraud identification. In this paper, the interpretability study is carried out from the three dimensions of object, feature, and time, and it is found that the four secondary indicators of solvency, ratio structure, governance structure, and internal control environment have a universal role in fraud identification, but the environmental indicators only have a greater role in certain industries that involve high pollution. The study also finds the logical differences between convolutional neural networks in determining non-fraud and fraud samples: non-fraud enterprises have similar features pattern, and the secondary indicators that play a major role for those companies are ratio structure, operating ability, social responsibility, and governance structure, while that of fraud enterprises varies greatly according to the type of fraud; the important features in all yeas of non-fraud samples have a role in the identification of fraud, while that of fraud samples often tend to be aggregated in a small time region. This paper analyzes the typical case, Guanong Shares, and the analysis results show that the secondary indicators that contribute the most to the determination of fraud of Guanong Shares in 2022 are cash flow analysis, social responsibility, governance structure, and per-share indicators, which are in line with its actual fraudulent means.

# 目录

# Contents

# 1 引言

## 1.1 研究背景

2024 年 1 月 5 日，中国证监会开出了新年的第一张罚单，指控瑞华会计师事务所，在审计康得新复合材料集团股份有限公司（下称康得新）2015 年至 2017 年年度的财务报表时，未能勤勉尽责，出具的报告存在虚假记载，被罚没共计约一千八百万元，敲响了新年财务舞弊的第一声警钟。

财务舞弊，是指企业为了获取利益，不遵循财务会计报告标准，有意识地歪曲反映企业经营情况，做出不实陈述的财务会计报告，从而误导信息使用者的决策。近年来，A 股和美股的舞弊案件频发。2012 年 A 股最严重的康美药业造假案被披露，康得新被先后发现涉嫌连续四年财务造假。上市公司财务舞弊严重损害了投资者利益、破坏了资本市场环境。

上市公司财务舞弊对全社会有着重大的负面影响，但财务舞弊的识别却十分困难。传统的舞弊识别方法多采用经验推理方法以及统计模型，这样的方式虽然可解释性较强，但很难捕捉多个特征之间非线性的相互作用。近年来，因为机器学习方法能够利用更大规模、更高维度的数据，捕捉更复杂的勾稽关系，使用机器学习方法识别上市公司财务舞弊的方式越来越受到学界关注。但机器学习方法的可解释力度较差，而财务舞弊的识别通常需要给出审计意见、指明判罚原因，所以机器学习方法在实务层面的应用较为受限。因此，如何更好地利用机器学习技术识别财务舞弊，并对其做出的决策进行解释，是目前机器学习模型在解决实际舞弊识别问题上所面临的新课题。

## 1.2 研究现状

### 1.2.1 财务舞弊识别相关研究综述

在机器学习的概念被提出之前，学者多采用逻辑回归等传统统计方法来识别

财务舞弊。近年来，机器学习方法显著地提升了财务舞弊预测问题的准确性。Liu 等人将随机森林模型与其他四种参数和非参数模型比较，得出随机森林模型的准确率最高[1]。周卫华等人首次使用 XGBoost 方法预测上市公司财务舞弊，并通过 Benford 定律、LOF 局部异常法、IF 无监督学习法，解决了灰色样本问题[2]。除了树模型之外，深度学习模型也被应用于财务舞弊的研究中。在二十世纪初，大多数研究采用全连接神经网络，近年来，部分学者考虑到年报文本信息存在的语义提示作用，循环神经网络（RNN）等能够处理序列数据的模型也逐渐被采纳[3]。特别地，除了在图像处理领域，卷积神经网络也被证实在处理非图像数据和序列数据方面也有着很好的表现。Alok Sharma 等人将非图像数据转换为图像数据，并利用卷积神经网络提取特征[4]；Zhao 等人将卷积神经网络框架用于时间序列数据的分类，其在分类精度和噪声容限方面均优于现有的时间序列分类方法[5]。

### 1.2.2 可解释性理论相关研究综述

虽然机器学习方法性能强大，但因其结构复杂、参数众多，在提升性能的同时牺牲了模型的可解释性，尤其是其中的深度学习方法，通常有上万甚至数十万个参数，可解释性很差，因此被称为"黑盒（black box）模型"。深度学习模型缺乏可解释性的缺点，导致其在高风险领域的应用受限[6]。

2018 年，Miller 首先对人工智能领域的可解释性（interpretability）做出了定义：观察者（人类）可以理解决策原因的程度[7]。同年 Christoph Molnar 发表了第一本系统性介绍机器学习可解释性的书籍《Interpretable Machine Learning》，并根据解释的范围分为全局（Global）可解释性和局部（Local）可解释性[8]。

一些解释方法适用性较广，如计算 Shaply 值来解释每个特征对于模型预测的贡献程度。由于深度学习模型具有特殊的隐藏层结构，及其训练过程中的梯度记录，我们可以利用这些特性探索专用于深度学习模型的解释方法，提高计算效率[8]。

目前，针对卷积神经网络的可解释性研究已经有了初步的进展。对于卷积神经网络来说，由于参数量级大、网络层数深，全局解释的难度很高。局部解释聚焦于解释某个有代表性的样本点或某组相似的样本簇，难度较低，解释效果较为

直观，如蒋家伟将 ICU 患者的生理文本数据转换为图片数据，通过迁移学习对患者进行生存、死亡预测，并使用梯度类激活映射图（graddient based class activate mapping，Grad-CAM）对其进行解释[9]；苏盈利用表征可视化技术和反卷积神经网络对电力系统负荷模式进行双向解释[10]；付贵山使用卷积神经网络对乳腺超声图像进行分类，并通过热度图和语义回归两种方法对模型进行可解释[11]。

### 1.2.3 文献述评

本文定义了两种财务舞弊识别模式：事后检测，即在企业当年年报数据公布后检测当年是否发生舞弊，这是传统统计方法以及无法处理序列数据的机器学习模型所采用的模式；事前预测，即提前一年预测下一年舞弊的可能性，这种模式需要模型具有处理序列数据的能力。

虽然现有文献已经显著提升了机器学习的识别准确率，但基本都是将财务舞弊数据当作一般的表格数据，实现事后检测。然而本文认为，财务舞弊数据是具有时间、对象、特征三个维度的面板数据，如果将财务舞弊数据当作普通格式化表格数据，则不能很好地利用时间维度的信息，难以实现事前预测；但如果分别为每个对象的多元时间序列数据构建专用于序列数据的模型，则会因为单个对象时间序列较短，无法满足机器学习的数据量要求。

此外，虽然现有文献已经提出了许多可解释机器学习方法，但对于财务舞弊识别的解释仍停留在特征重要性排序上，解释力度较弱，无法提供特征、时间、对象三个维度的综合解释，对于财务舞弊识别的指导意义不足。

## 1.3 研究目的及意义

本文旨在实现财务舞弊的事前预测，并解释黑盒模型的内部决策逻辑。基于现有文献存在的问题，同时受康得新连续多年财务造假的典型案例的启发，本文提出利用卷积神经网络，充分提取财务舞弊面板数据时间、特征两个维度的信息，并探索了针对于面板数据的"类图像"处理方式，实现了财务舞弊事前预测。此外，针对"黑盒模型"可解释性差的特点，本文采用专用于卷积神经网络的可视化解释方案，使用局部可解释方法分析不同种类样本的特征。

本文提出的方法具有理论和实际意义。在理论上比较了不同模型的优劣，并基于卷积神经网络提出了一种可以充分利用面板数据信息、适用于大数据背景的财务舞弊识别模型；在实践中可以为审计人员提供辅助，提示可能的舞弊路径，节省人工查证时间、提高舞弊检出效率，同时可以为投资者投资行为做出预警，对未来人工智能融入资本市场运作有着较为重要的意义。

## 1.4 论文组织结构

本论文共分为六章，论文首先分析了传统财务舞弊识别方法以及机器学习方法存在的时效性差、解释难等问题，接着总结了已有文献研究，并基于此提取本文研究的目的及意义。之后论文回顾了理论方法，建立了基于卷积神经网络的财务舞弊识别模型并构建了指标体系，并选取中国 A 股企业的多维数据做实证研究，并分析具有代表性的案例，给出人工智能在审计实务中可能的发展方向以及实操方案。

论文具体安排如下：

第 1 章 提出了论文的研究背景、回溯了现有文献，并基于现有研究的不足提出本文的目标及意义。

第 2 章 回顾理论方法，介绍了卷积神经网络的发展历程、层级结构。

第 3 章 基于历史上两种成功的卷积神经网络结构，构建了适用于本文任务的网络结构，并介绍了指标体系构建的依据、标准及具体内容。

第 4 章 使用第 3 章构建的神经网络进行实证研究，介绍了本文的模型评价指标，进行参数调优，并使用其他两种模型进行模型比较。

第 5 章 采用反向解释、前向解释相结合的方式，对第 4 章的实证结果进行可解释性分析，并分析有代表性的案例。

第 6 章 总结了论文所做的工作，指明了下一步的改进计划：主要考虑改变网络结构，使其与自然语言处理模型融合，以图像、语义结合的解释方式进一步提升网络可信度与用户体验。

# 2 相关理论介绍

## 2.1 卷积神经网络概述

卷积神经网络（Convolutional Neural Networks，CNN）是一类起步较晚，但发展迅速、应用广泛、性能强大的深度学习模型，主要利用空间不变性，通过使用较少的参数，来处理网格化数据，如图像、语音和文本等，达到较好的效果。

最早的 CNN 可以追溯到 1989 年，由 AT&T 贝尔实验室的研究员 Yann LeCun 提出的 LeNet 模型，目的是识别图像中的手写数字，其含有卷积层、池化层、全连接层等基本组件，在小型数据集上达到了与支持向量机相当的结果。2012 年，Alex Krizhevsky 等人基于 LeNet 框架构建了更深、更复杂的网络，并做了可以使模型在 GPU 上训练的改进。同时将激活函数从 sigmoid 更换为 ReLU，ReLU 激活函数的计算更简单，不需要复杂的求幂运算，使模型能更好地处理大规模数据集。2015 年，何恺明等人用残差块（residual blocks）使输入可以通过层间的残余连接更快地向前传播，提出了残差网络（ResNet）。至今，CNN 已经广泛应用于计算机视觉、自然语言处理，以及时间序列分析等领域。

## 2.2 卷积神经网络结构

卷积神经网络的主要组成分为：输入层（Input Layer）、卷积层（Convolutional Layer）、池化层（Pooling Layer）、全连接层（Full Connection Layer）。其中，卷积层和池化层是卷积神经网络所特有的。

卷积层是卷积神经网络最核心的组成部分，通过参数共享机制提取图像的局部信息。参数共享机制，指的是在卷积层中，输入张量的某一局部，与相同维度和大小的卷积核张量，通过互相关运算产生输出张量中的一个元素，如图 2-1 所示。

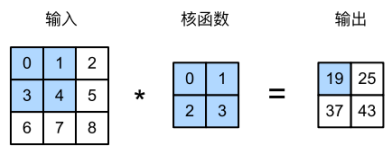

**图 2-1 二维互相关运算示意图**[12]

互相关即矩阵对应元素相乘后相加，图中蓝色部分所示计算为 0×0 + 1×1 + 3×2 + 4×3 = 19

为了更好提取图像边缘像素的信息，卷积层引入了填充（padding）的概念。在输入图像的边界填充元素即为 padding，填充的元素通常为 0。在计算互相关时，卷积窗口从输入张量的左上角开始，向下、向右滑动，每次滑动元素的数量即为步幅（stride）。

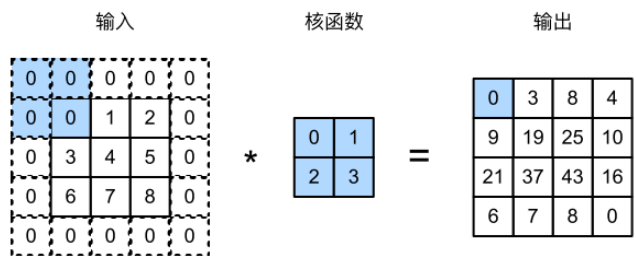

**图 2-2 带填充的二维互相关示意图**[12]

原始矩阵四周虚线方格即为填充部分，图中蓝色部分所示计算为 0×0 + 0×1 + 0×2 + 0×3 = 0，向右滑动一格后的计算为 0×0 + 0×1 + 0×2 + 1×3 = 3，向下滑动一格后的计算为 0×0 + 0×1 + 0×2 + 3×3 = 9

池化层，又叫汇聚层，通常设置在卷积层之后，通过逐渐聚合卷积层提取的局部信息，压缩中间层尺寸，提取特征关键信息，最终实现全局学习，通常分为最大池化层和平均池化层，其中最大池化层的应用较为广泛，其工作原理如图 2-3 所示。

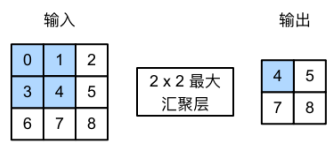

**图 2-3 最大汇聚层工作原理**[12]

# 3 模型建立

## 3.1 神经网络架构

本文在 LeNet 的基础上，参考 AlexNet 对 LeNet 的改进思路，构造了适合于财务舞弊识别的卷积神经网络。本文构建了适用于事后检测、事前预测两种识别模式的网络，这两种模式的识别逻辑基本一致，但识别难度有所差异，因此两张网络的基本结构及初始参数相同，如图 3-1 所示，但参数调节不同，见 4.3.3 节。

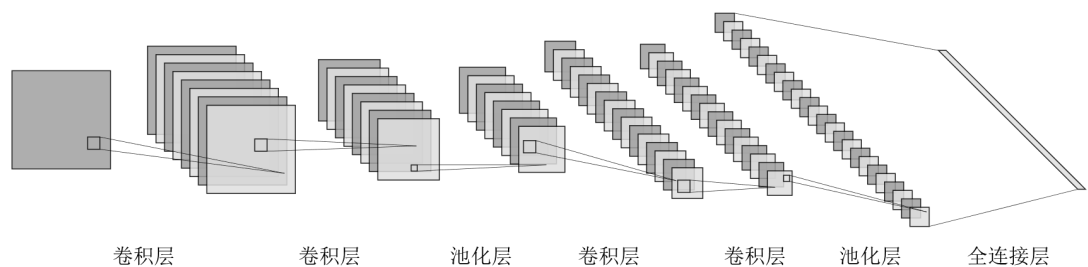

卷积层　　卷积层　　池化层　　卷积层　　卷积层　　池化层　　全连接层

**图 3-1 CNN 网络架构示意图**

依图 3-1 所示，网络的主体部分由两组相似的层级结构组成：卷积层-卷积层-池化层-dropout 层。相比卷积层和池化层交替连接的结构来说，两个卷积层的堆叠可以让网络逐渐提取更加抽象和复杂的特征；池化层可以帮助减小特征图的尺寸，降低模型的计算复杂度。由于灰度图维度较小，卷积核的尺寸设置为 3*3，随着网络加深，卷积核的个数从 32 个增大到 64 个，并在卷积图四周做填充，确保网络在压缩图片尺寸的时候不损失图片的信息量。dropout 层防止了模型过拟合，由于全连接层参数多于卷积层和池化层，因此将卷积池化层的 dropout 概率设置为 0.25，全连接层的 dropout 概率设置为 0.5。相比于 LeNet 采用的 sigmoid 激活函数，本文在卷积层采用了 AlexNet 的 ReLU 函数作为激活函数，ReLU 函数是一种常见的非饱和函数，有效地避免了梯度消失问题。

在经过卷积池化操作之后，网络设置了一层扁平层，用于将网络训练过程中的三维数据展平成全连接神经网络可以接收的一维数据。接着由全连接层进行二

分类概率的预测，全连接层的激活函数选择了二分类任务所需要的 sigmoid 函数。

表 3-1 神经网络初始超参数

| 损失函数 | 优化器 | 学习率 | batch_size | epoch |
| --- | --- | --- | --- | --- |
| binary_crossentropy | Adam | 0.01 | 64 | 5 |

网络的其他初始超参数如表 3-1 所示。本文采用常用的二分类问题损失函数"二元交叉熵（binary cross entropy）"，其计算公式为：

$$Loss = -\frac{1}{N}\sum_{i=1}^{N} y_i \cdot log(p(y_i = 1)) + (1 - y_i) \cdot log(1 - p(y_i = 1)) \quad (3-1)$$

其中$y_i$是第$i$个样本的二元标签值 0 或者 1，$p(y_i = 1)$是模型对第$i$个样本的预测值，即模型预测第$i$个样本标签值为 1 的概率。对于标签为 1 的情况，如果预测值$p(y_i)$趋近于 1，二元交叉熵趋近于 0，表明模型分类效果较好；反之，如果此时预测值$p(y_i)$趋近于 0，二元交叉熵趋近于 1，表明模型分类效果较差。

本文采用 Adam 优化器作为网络优化器，Adam 算法使用自适应学习率，可以根据每个参数的梯度情况自动调整学习率，同时引入了动量机制，类似于其他优化器中的动量项，收敛速度更快，且不易陷入局部最优解，是目前最有效、应用最广泛的优化器。

## 3.2 财务舞弊识别指标体系构建

指标体系构建对于机器学习模型来说至关重要。梁力军等人构建了财务指标体系，但仅有财务指标不足以把握公司运作过程中的复杂逻辑[13]。为此，叶钦华等人基于复式簿记与会计信息系统论，除财务税务指标维度外，还考虑了公司治理（Govern）等其他四个维度，五个维度分别对应于会计信息生产的各个环节[14]。但随着可持续发展理念深入人心，环境（Environment）与社会责任（Social）因素对于企业风险预测的作用越来越显著[15-16]。因此，本文参考叶钦华等人提出的"五维度"指标体系，并做增删，构建了含有财务指标、ESG 指标、内部控制指标的三级指标体系，较完整地反应了企业的经营情况。

在构建指标体系时，本文秉承以下三点原则：其一，由于模型可解释性的前

提是特征（feature）具有解释性[8]，所以指标体系中的每个特征都必须具有现实意义；其二，对于财务指标，由于不同公司规模不同、业务不同，所有应该尽量构造相对指标，尽量避免绝对指标；其三，对于 ESG、内部控制指标，由于不同公司内部结构不同、不同环境指标单位不同，所以应尽量构造定性指标，尽量避免定量指标。基于此，本文构建的财务舞弊识别指标体系如表 3-2 所示。

表 3-2 财务舞弊识别指标体系

| 一级指标 | 二级指标 | 三级指标 |
| --- | --- | --- |
| 财务指标 | 偿债能力 | 流动比率、速动比率、现金比率等 |
| | 披露财务指标 | 非经常性损益等 |
| | 比率结构 | 流动资产比率、现金资产比率等 |
| | 经营能力 | 应收账款与收入比等 |
| | 盈利能力 | 资产报酬率、总资产净利润率等 |
| | 现金流分析 | 净利润现金净含量等 |
| | 风险水平 | 财务杠杆、经营杠杆等 |
| | 发展能力 | 资本保值增值率等 |
| | 每股指标 | 每股收益等 |
| | 相对价值指标 | 市盈率（PE）等 |
| | 股利分配 | 股利分配率等 |
| ESG指标 | 环境披露 | 废气减排治理情况等 |
| | 社会责任 | 是否经第三方机构审验等 |
| | 治理结构 | 前十大股东是否存在关联等 |
| 内部控制指标 | 内部控制环境 | 独立董事比例等 |
| | 风险管理 | 是否披露经营管理风险等 |
| | 控制活动 | 存货是否有损失等 |
| | 信息与沟通 | 是否有机制管理与投资者的关系等 |
| | 监督活动 | 是否有内部控制评价报告等 |

# 4 实证研究

## 4.1 样本选择

本文的原始数据来自于 CSMAR 数据库的公司研究系列，以及 CRNDS 数据服务平台的内部控制研究系列，时间跨度为 2010~2022 年，共三万余条样本、326 个特征。

舞弊样本的认定基于 CSMAR 数据库《违规信息总表》中记载的，在上海证券交易所和深圳证券交易所上市、并被证券监督管理委员会公开披露的违规记录。由于上市公司财务舞弊的手段主要是对利润和资产做出虚假的记载或披露，本文选择 CSMAR 数据库所有违规类型中的四类作为财务舞弊样本，分别为"P2501 虚构利润""P2502 虚列资产""P2503 虚假记载（误导性陈述）""P2506 披露不实（其它）"[17]。样本标签列名为"是否舞弊"，舞弊样本标记为 1，作为正样本，非舞弊样本标记为 0，作为负样本。

## 4.2 特征工程

### 4.2.1 缺失值处理

原始数据集中共有 326 个特征，其中有 41 个特征的缺失值超过了 30%，其余特征的缺失值不足 5%。第一步，本文先手动剔除了缺失值超过 30%的特征。第二步，考虑某个股票标的所有年份都缺失某一特征的情况极少出现，因此，本文利用同一股票标的同一指标的相似性，同时考虑到财务数据和非财务数据的差异性，根据股票代码对数据集进行分类后，对缺失值不足 5%的财务特征采用插值填充法，非财务特征采用 KNN 填充法。第三步，本文删除了所有年份都缺失某一特征的样本。缺失值处理后的数据集共 27650 个样本，283 个特征。

### 4.2.2 灰色样本剔除

灰色样本是指由于证监会认定财务舞弊存在滞后性或虽然实际舞弊但尚未发现和披露的样本[2]。考虑到本文采用的数据集样本多、维度高，本文采用孤立森林（Isolation Forest）算法剔除灰色样本[18]。孤立森林方法是一种基于树模型的异常值检测方法，通过递归构造一系列随机生成的二叉树来识别异常值所在的区域。该方法的基本思想是：相对于正常值来说，异常值通常存在于在特征空间中密度较低的位置，因此可以通过路径较短的二叉树将其分隔出来。孤立森林算法共分为两步，第一步是训练过程，第二步是评分过程，某个样本$x$的异常度分数$s$定义为：

$$s(x, \varphi) = 2^{-\frac{E(h(x))}{c(\varphi)}} \tag{4-1}$$

其中$\varphi$为训练样本个数，$E(h(x))$表示样本$x$在所有并行二叉树上的路径长度均值，$c(\varphi)$为正则项系数，定义为：

$$c(\psi) = \begin{cases} 2H(\psi-1) - 2(\psi-1)/n & \text{for } \psi > 2, \\ 1 & \text{for } \psi = 2, \\ 0 & \text{otherwise.} \end{cases} \tag{4-2}$$

其中，$H(i)$为谐波级次，可用欧拉常数估计。

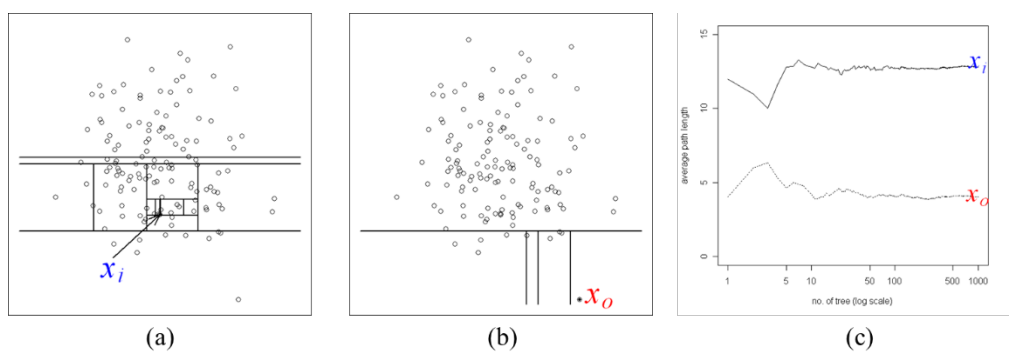

**图 4-1 孤立随机森林算法示意**[18]
**(a)正常样本分割示意图;(b)异常样本分割示意图;**
**(c)正常样本与异常样本的分割二叉树长度对比**

### 4.2.3 Z-Score 标准化

剔除灰色样本后，由于各公司规模不同、各特征单位不同，为了防止神经网络梯度爆炸，本文采用 Z-Score 方法对样本进行标准化处理，公式如下：

$$z = \frac{x-\mu}{\sigma} \tag{4-3}$$

其中，$z$ 为标准化之后的样本，$x$ 为原样本，$\mu$ 为样本集均值，$\sigma$ 为样本集标准差。

### 4.2.4 "类图像"转换

本文创新性地将面板数据做分割，以满足卷积神经网络的输入数据维度要求。之所以将这一过程命名为"类图像"转换，是因为本文只改变了数据的维度，以利用卷积神经网络突破传统舞弊识别的事后检测、实现事前预测，并没有改变数据的存储格式，样本仍以数据帧的格式存储。

以事前预测网络为例，本文首先将有 2022 年数据的公司挑选出来，并提取出时间序列长度大于等于 6 年的公司共 1436 家。筛选后的面板数据维度为（1436，12，283），维度从左至右依次为样本数量、时间跨度、特征个数。随后根据"股票代码"分割面板数据，将 1436 家公司 2010~2021 年的多元时间序列数据，视作 1436 张二维灰度图像，其中，将数据帧的时间维度视作图像的宽，特征维度视作图像的长，并以该公司 2022 年是否舞弊作为图像标签。由于卷积神经网络对图像像素的相对位置敏感[4]，本文在将表格数据视作灰度图像时，对于图像的长（数据帧的特征维度），严格按照财务指标、ESG 指标、内部控制指标从左向右依次排列。

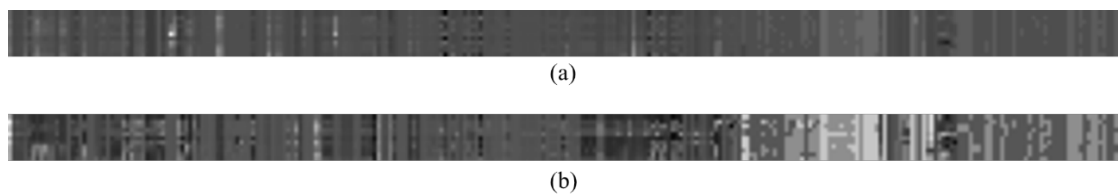

**图 4-2 分割后的单个样本二维灰度示意图**
**(a)深振业 A(股票代码:000006); (b)西陇科学(股票代码:002584)**

**4.2.5 SMOTE 过采样**

经过"类图像"转换的 1436 个公司实例中，正常样本有 1367 家，舞弊样本有 69 家，仅占全部样本的 4.8%，因此本文在投入网络训练前，使用 SMOTE（Synthetic Minority Over-sampling Technique）方法做了过采样处理。

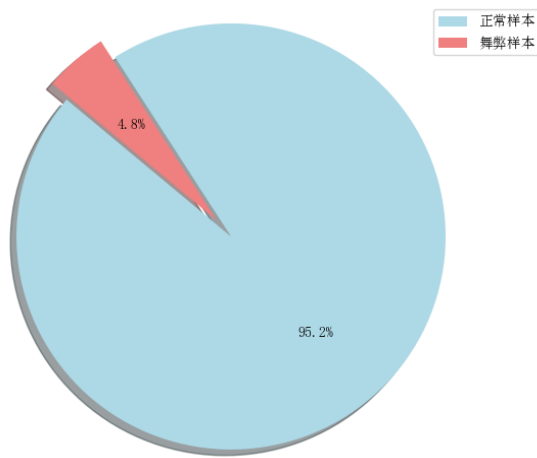

**图 4-3 数据处理后的样本分布图**

SMOTE 通过在少数类样本附近随机插值实现对少数类样本的扩充，按照如下的公式构建新的样本：

$$x_{new} = x + rand(0，1) \times (\tilde{x} - x) \tag{4-4}$$

其中，$x_{new}$ 为新样本，$x$ 为少数类样本的原样本，$rand(0，1)$ 表示符合均匀分布的随机数，$\tilde{x}$ 表示少数类样本的 k 近邻。本文先将处理后的二维灰度图像展平为一维向量，使用 SMOTE 算法对向量进行过采样处理之后，再还原成灰度图像作为模型的输入。过采样后正负样本各 1367 个。

## 4.3 模型训练与比较分析

**4.3.1 样本划分**

本文按照标签"是否舞弊"将数据集按照训练集：验证集：测试集 = 70%：15%：15%的比例划分，划分后的样本分布如表 4-1 所示，由于本文将单个样本

的数据视为二维灰度图像，所以最后一个维度"通道数"为 1。

表 4-1 神经网络样本划分

|  | 训练集 | 验证集 | 测试集 |
|---|---|---|---|
| 正样本 | 957 | 205 | 205 |
| 负样本 | 956 | 205 | 206 |
| 事后检测样本维度 | (1913, 13, 283, 1) | (410, 13, 283, 1) | (411, 13, 283, 1) |
| 事前预测样本维度 | (1913, 12, 283, 1) | (410, 12, 283, 1) | (411, 12, 283, 1) |

### 4.3.2 模型评价指标

即使经过采样处理，本文原始样本中存在的数据不平衡问题也会对模型的预测结果产生一定影响，此外，将舞弊样本错分为正常样本的代价要远大于将正常样本错分为舞弊样本。因此，本文认为准确率（accuracy）不适用财务舞弊的检测任务。

本文在模型训练阶段采用 AUC（Area Under the ROC Curve）作为评价指标。AUC 是一种用于评估二分类模型性能的指标，计算了 ROC（Receiver Operating Characteristic Curve）曲线下的面积，取值范围为[0，1]，越接近 1 表示模型性能越好。AUC 在反映正样本的分类准确率的同时，考虑了将负样本错分为正样本的代价。

本文在模型比较阶段采用 AUC、召回率（recall）、F2-score 三个指标进行综合评价。召回率的计算公式为：

$$Recall = \frac{True\ Positive}{True\ Positive + False\ Negative} \quad (4\text{-}5)$$

表示样本中正例预测正确的比例。F2-score 是 F beta-score 中 beta 取 2 的特例，F beta-score 的计算公式为：

$$F\ beta - Score = (1 + \beta^2) \cdot \frac{Precision \cdot Recall}{\beta^2 \cdot Precision + Recall} \quad (4\text{-}6)$$

当 $\beta > 1$ 时，认为 recall 更重要，当 $\beta < 1$ 时，认为 precision 更重要，本文的分类任务中正样本更重要，因此选择 $\beta = 2$。

### 4.3.3 超参数调节

本文需要调节的超参数为学习率、batchsize、epoch、分类阈值。考虑到学习率与 epoch 具有反向相关性，当学习率较大时，梯度下降速度较快，因此达到最优所需要的 epoch 较少，反之亦然[19]。由于事后检测与事前预测在逻辑和难度上存在一定差异，两个网络的参数应该有所区分。因此本文调节参数的顺序为：先同时调整学习率和 epoch，再分别调节 batchsize 和分类阈值；先调节事后检测网络，再按照相同步骤调节事前预测网络。

首先，本文成比例尝试了不同的学习率与 epoch 组合，事后检测网络训练过程中验证集的 AUC 如图 4-4 所示。

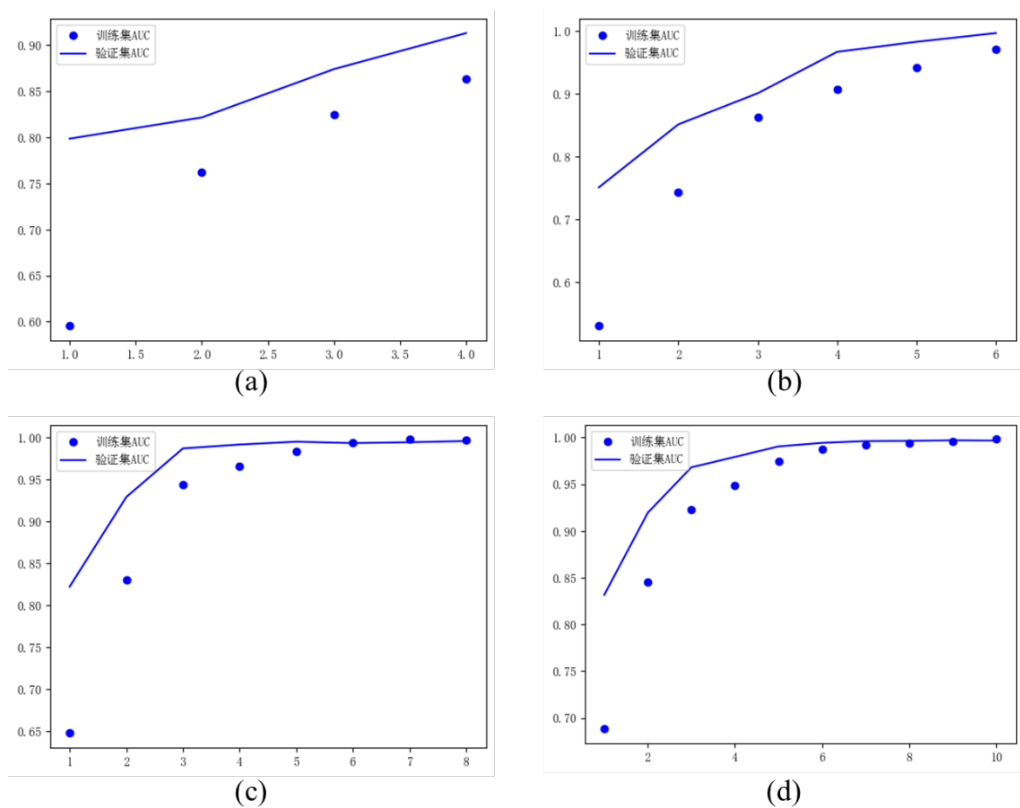

**图 4-4 事后检测网络训练效果随学习率、epoch 变化图**
**(a)学习率=0.01, epoch=4; (b)学习率=0.005, epoch=6**
**(c)学习率=0.001, epoch=8; (b)学习率=0.0005, epoch=10**

由图 4-4 可知，前两种参数组合的训练集与验证集曲线未出现交点，说明模型欠拟合，最后一种参数组合的训练集与验证集曲线出现交点后又分离，说明模型过拟合。同时观察到图 4-4(c)在第 6 个 epoch 后出现分离趋势，因此对 epoch 做微调,最终确定事后检测网络的最优学习率和 epoch 组合为 learning_rate=0.001，epoch=6。

BatchSize 是卷积神经网络一次训练的样本数目，因为训练样本较多，而计算机内存有限，因此需要 BatchSize 来指定每次训练的数量。BatchSize 太小容易因为每一批次样本代表性不足而带来较大的方差，使得模型不稳定，收敛速度也较慢；BatchSize 太大容易因为每一批次样本数量过多，不利于模型跳出局部最优解，也可能导致模型泛化能力较差。不同 BatchSize 对事后检测网络的影响如图 4-5 所示。

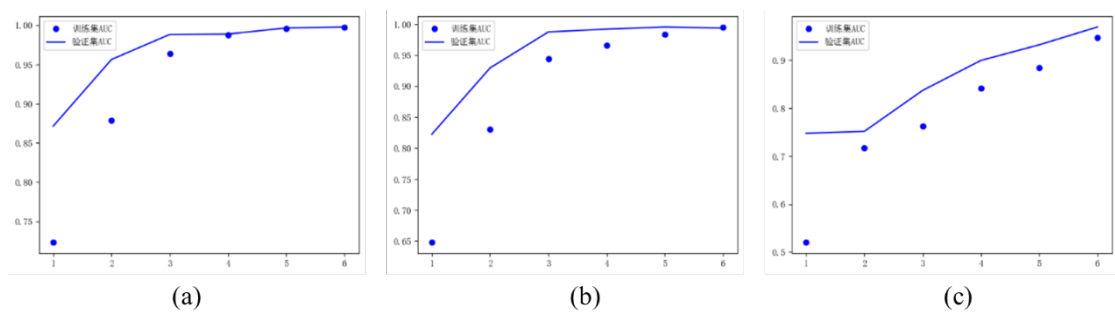

**图 4-5 事后检测网络训练效果随学习率、BatchSize 变化图**
(a) BatchSize=32; (b) BatchSize=64; (c) BatchSize=128

由图 4-5 可知，当 BatchSize=32 时模型拟合良好，当 BatchSize=64 时模型过拟合，当BatchSize=128时模型欠拟合。因此,事后检测网络的最优BatchSize=32。

事前预测网络按照同样的顺序与方式调节,除了卷积神经网络训练过程中比较常见的超参数，由于本文任务的特殊性，超参数中的分类阈值需要特别进行调节。财务舞弊判定一旦出错，后果将非常严重，如果阈值太高，将舞弊公司判定为正常公司会造成投资者损失、影响上下游企业；如果阈值太低，将正常公司判定为舞弊公司，将会影响公司声誉、打击市场信心。因此，采用默认分类阈值 0.5 是不理智的。

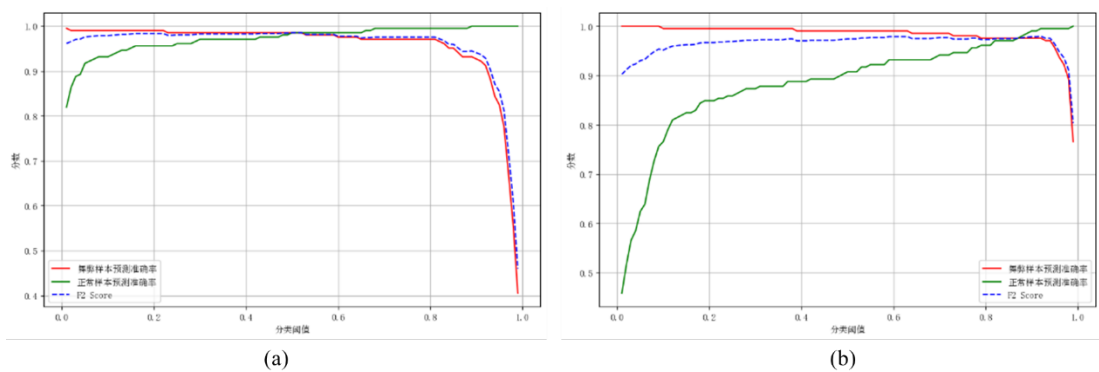

**图 4-6 神经网络训练效果随分类阈值变化图**
**(a) 事后检测网络阈值变化情况; (b) 事前预测网络阈值变化情况**

图 4-6 展示了事后检测和事前预测两个网络在不同分类阈值下的表现。我们可以从图中发现，事前预测网络的阈值分数图整体比事后检测网络右偏，这是因为事前预测的可用信息量比事后检测少一年，不确定性更强，因此在同样的分类阈值下，为了确保舞弊样本的预测准确率，会将更多的正常样本错分为舞弊样本。

考虑到财务舞弊判定的审慎性，在不误判过多正常样本的条件下提高舞弊样本检出率，本文将事后检测的最优分类阈值确定为 0.45，将事前预测的最优分类阈值确定为 0.75。

经过上述的调参过程，事后检测网络和事前预测网络的最优参数如表 4-2 所示，因为事后检测与事前预测的逻辑相似、难度不同，所以两个模型仅在分类阈值上有所区分。模型的最终训练效果如图 4-7 所示，表明模型拟合良好，无过拟合现象。

**表 4-2 神经网络最优参数**

| 预测模式 | learning_rate | epoch | BatchSize | 分类阈值 |
|---|---|---|---|---|
| 事后检测 | 0.001 | 6 | 32 | 0.45 |
| 事前预测 | 0.0005 | 8 | 64 | 0.75 |

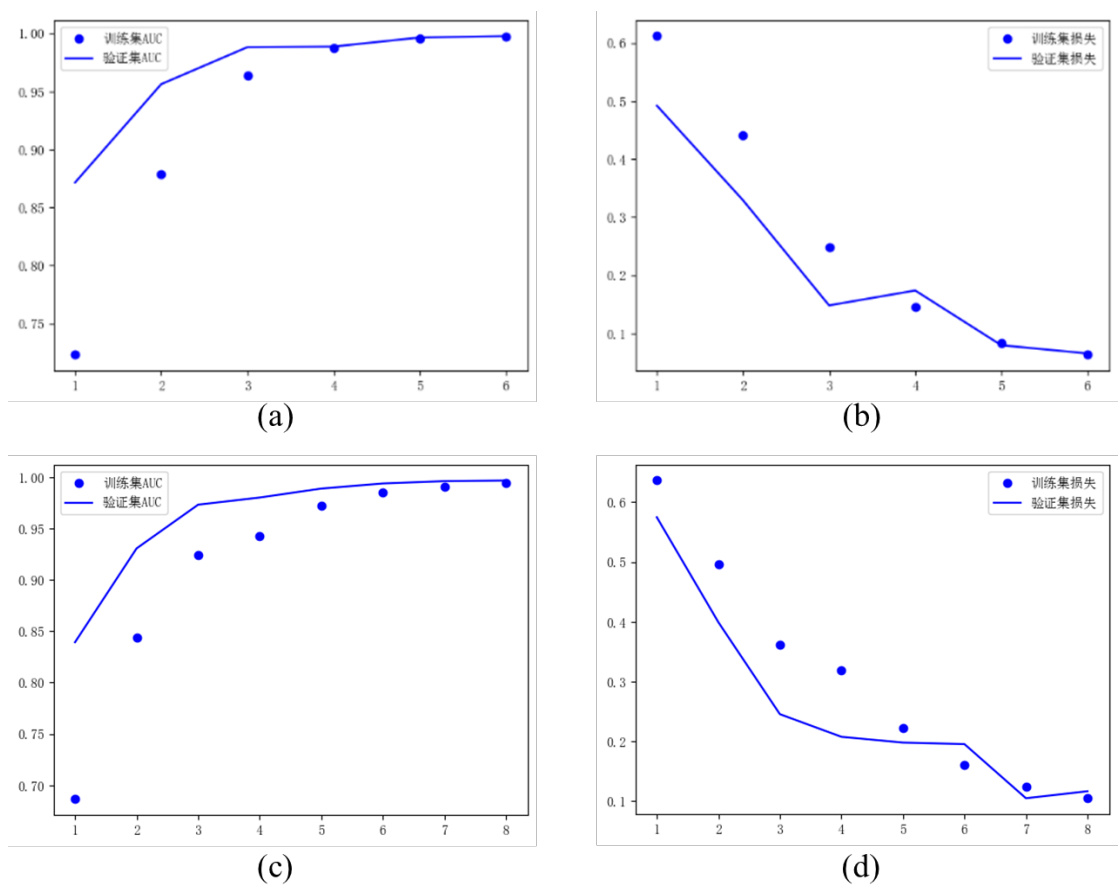

**图 4-7 事后检测与事前预测网络最终训练效果图**
**(a)事后检测网络 AUC; (b)事后检测网络损失**
**(c)事前预测网络 AUC; (b)事前预测网络损失**

本文选择 15%的样本作为测试集，事后、事前预测网络在测试集的表现如表 4-3 所示。因为难度更高、不确定性更强，事前预测网络的 AUC、F2-Score 略低于事后检测网络，但差别不大，表明卷积神经网络在事前预测上也有较强的可靠性。此外，事前预测网络为了维持舞弊样本的检出率，牺牲了一定的正常样本的预测准确率。

表 4-3 卷积神经网络在预测集上的表现

| 网络模式 | AUC | Recall | F2-Score | 舞弊样本准确率 | 正常样本准确率 |
|---|---|---|---|---|---|
| 事后检测 | 0.9929 | 0.9951 | 0.9893 | 99.5% | 96.6% |
| 事前预测 | 0.9875 | 0.9951 | 0.9808 | 99.5% | 92.2% |

### 4.3.4 模型比较分析

本文选用了逻辑回归模型，以及 LightGBM 模型作为对比模型，比较不同模型在处理财务舞弊识别问题上的表现。逻辑回归模型是使用统计方法以来，适用范围最广的财务舞弊识别方法，LightGBM 模型是近年来新兴的机器学习模型，以运算速度快、分类准确率高，尤其是擅长处理表格数据著名。

由于这两种模型不具有处理时间序列数据的功能，本文选取样本集中 2010~2017 年的数据作为训练集，2018~2019 年的数据作为验证集，2020~2021 年的数据作为测试集。

考虑到数据集的特征较多，本文为逻辑回归添加了 L1 惩罚项，计算公式如下：

$$J(\theta)_{L1} = C * J(\theta) + \sum_{j=1}^{n} |\theta_j|(j \geq 1) \quad (4\text{-}5)$$

其中，$J(\theta)$ 为损失函数，$C$ 是用来控制正则化程度的超参数，$n$ 是特征的总数，$\theta_j$ 代表每个参数。相比 L2 惩罚项，L1 惩罚项能迅速使对模型贡献不大的特征系数收缩到零，起到了特征选择的作用，更适用本文的数据集。逻辑回归的模型表现随分类阈值的变化情况如图 4-8 所示，本文选择 0.35 作为逻辑回归模型的最优分类阈值，逻辑回归模型最终参数及超参数设置如表 4-4 所示。

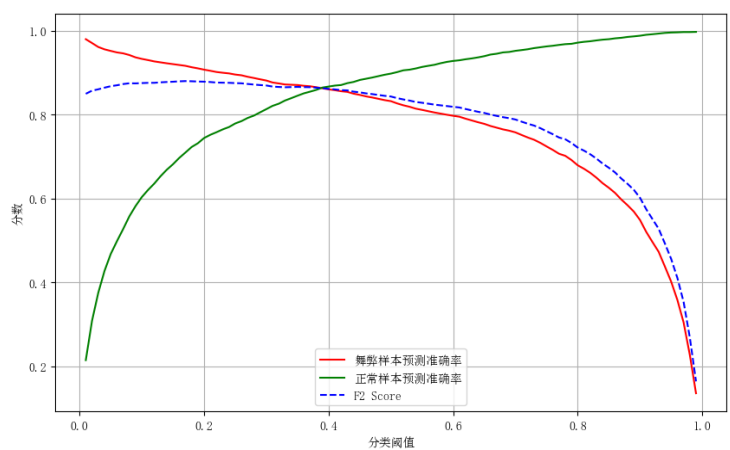

图 4-8 逻辑回归训练效果随分类阈值变化图

表 4-4 逻辑回归最优参数

| 模型名称 | 惩罚项 | solver | 分类阈值 |
| --- | --- | --- | --- |
| 逻辑回归 | L1 | Linear | 0.35 |

LightGBM（Light Gradient Boosting Machine）是一个基于梯度提升算法的集成模型，通过迭代地训练多个决策树模型，将所有决策树的结果通过加权集成输出最终结果。它采用直方图算法和按层生长的方法，在构建决策树和进行分裂时可以更快地处理数据，同时支持并行计算，在提高模型准确性的同时大幅度降低了训练时间。

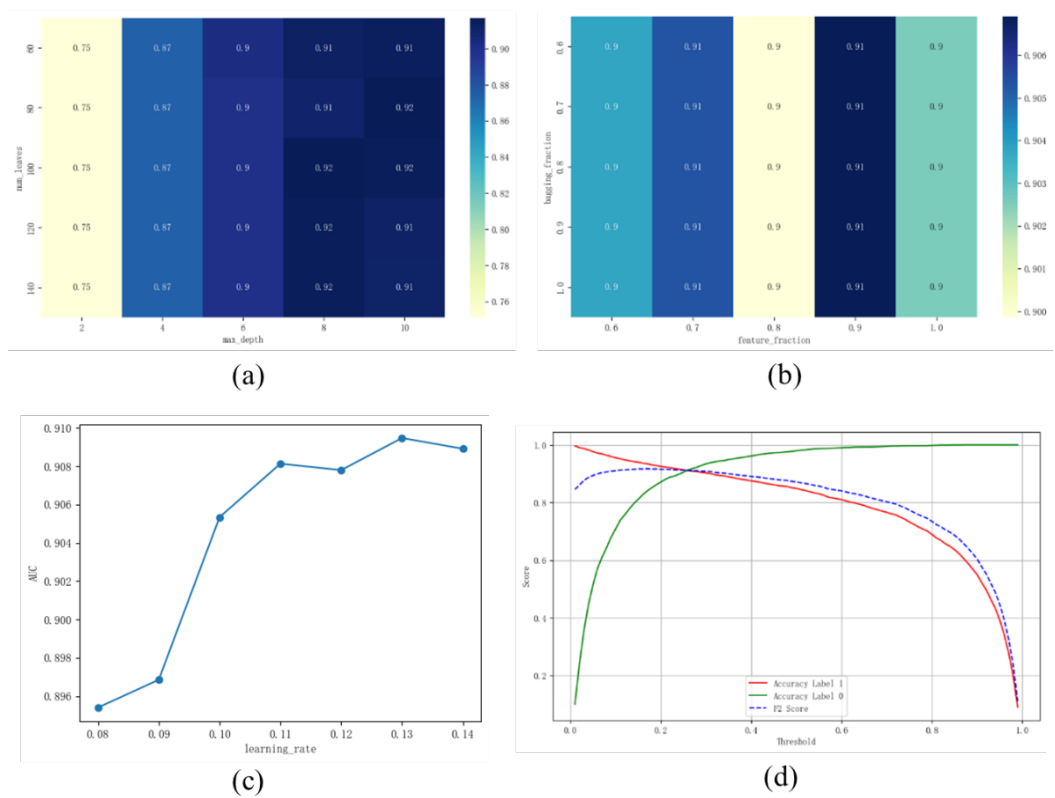

图 4-9 LightGBM 训练效果随参数变化图
(a) AUC 随 num_leaves, max_depth 变化图；
(b) AUC 随 bagging_fraction, feature_fraction 变化图；
(c) AUC 随学习率变化图; (d) AUC 随分类阈值变化图

本文对于 LightGBM 的调参策略为：先固定初始学习率为 0.1、分类阈值为 0.25，再使用网格搜索法，同时调节控制模型深度和广度的 num_leaves、max_depth，以及防止模型过拟合的 bagging_fraction、feature_fraction 两组参数，最后再来调整学习率和分类阈值。LightGBM 模型的参数搜索情况如图 4-9 所示，最终参数设置如表 4-5 所示。

表 4-5 LightGBM 最优参数

| 模型名称 | num_leaves | max_depth | bagging_fraction | feature_fraction | learning_rate | 分类阈值 |
| --- | --- | --- | --- | --- | --- | --- |
| LightGBM | 100 | 6 | 0.7 | 0.7 | 0.13 | 0.2 |

卷积神经网络、逻辑回归、LightGBM 模型在测试集上预测的结果如表 4-6 所示。可以看到，卷积神经网络的预测性能明显优于逻辑回归以及 LightGBM，而且事前预测的卷积神经网络也并没有因为更少的信息量和更大的不确定性，损失过多的性能，表明卷积神经网络在财务舞弊识别问题上的强大性能以及鲁棒性。

表 4-6 各模型在预测集上的表现

| 模型名称 | AUC | Recall | F2-Score | 舞弊样本准确率 | 正常样本准确率 |
| --- | --- | --- | --- | --- | --- |
| 逻辑回归 | 0.8447 | 0.8438 | 0.8441 | 84.4% | 84.6% |
| LightGBM | 0.8929 | 0.9161 | 0.9077 | 91.6% | 87.0% |
| 事后检测 CNN | 0.9929 | 0.9951 | 0.9893 | 99.5% | 96.6% |
| 事前预测 CNN | 0.9875 | 0.9951 | 0.9808 | 99.5% | 92.2% |

各模型输出的测试集预测概率的分布直方图如图 4-10 所示，可以清晰直观地看出，卷积神经网络输出预测概率的直方图两侧更集中、中间更稀疏，表明卷积神经网络对于舞弊样本和正常样本的区分更明显。

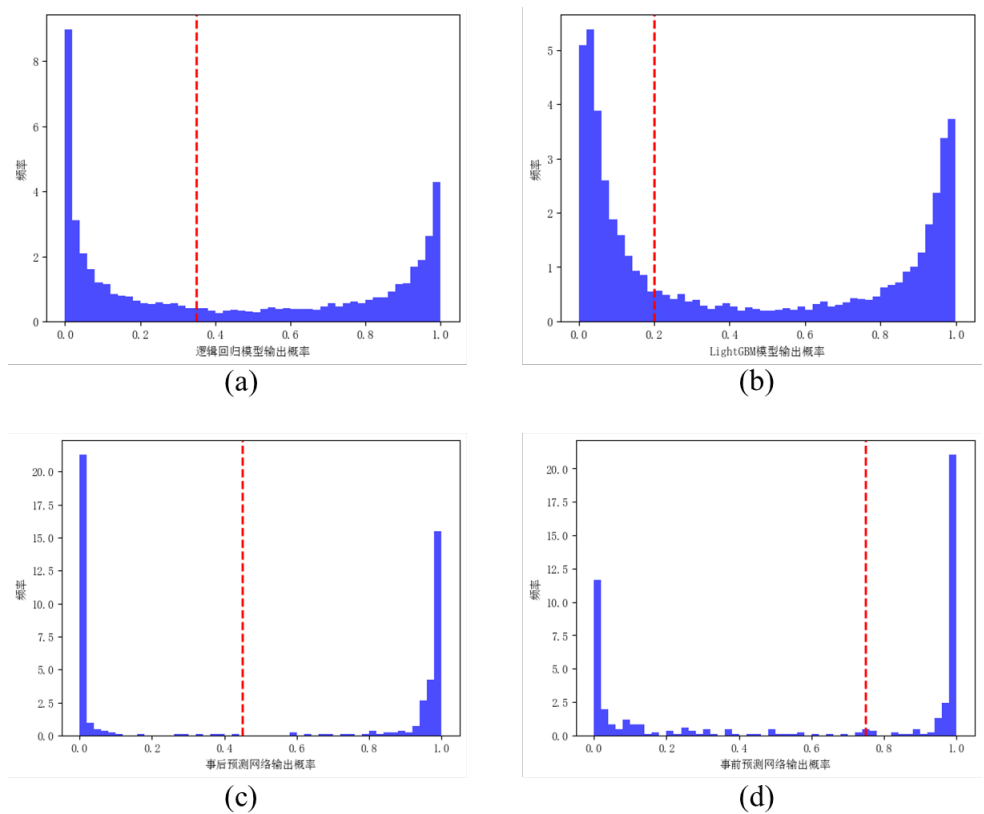

图 **4-10** 各模型输出预测集概率直方图

图中红色虚线为各模型最优分类阈值

**(a)**逻辑回归输出概率直方图; (b) LightGBM 输出概率直方图;

**(c)**事后检测网络输出概率直方图; **(d)** 事前预测网络输出概率直方图

基于卷积神经网络的上市公司财务舞弊识别及可解释性研究

# 5 模型可解释性分析

在财务舞弊识别情景中，审计师和会计师经过上百年的努力，已经建立了完善的审计流程和成熟的舞弊原因溯源机制，并且由于财务舞弊识别的审慎性，在现阶段尚且不能依靠深度学习输出的概率值就完成对某个公司的舞弊识别。因此，在现阶段深度学习的主要任务是对人工检测进行事前预警，提示审计师某个企业可能的舞弊路径，并在此基础上进行有针对性的取证调查。

基于以上两点考虑，本文采用表征可视化方法，提取卷积神经网络在前向传播过程中的卷积层动态变化，来反映神经网络学习的过程；采用梯度类激活映射图方法，来反映神经网络学习的结果。梯度类激活映射图通过计算 CNN 目标类别相对于最后一个卷积层输出特征图的梯度，按照特征图中的空间位置进行加权平均，进行激活、插值并拉伸覆盖到原始图像上，表示图像各部分对于预测结果的影响。

首先，本文通过分析不同类型样本的卷积层特征图、梯度类激活映射图，来总结神经网络的学习过程和决策逻辑，说明卷积神经网络的可靠性；接着，选取典型的舞弊样本进行案例分析，说明卷积神经网络的有效性，并提出本文所构建的网络在审计实务中的使用建议。

## 5.1 黑盒模型决策逻辑分析

本文根据会计准则对于财务舞弊的分类，选取 2022 年不同舞弊类型、来自不同行业的代表企业，基于事前预测网络，提取其 2010~2021 年数据在最后一层卷积层的梯度类激活映射图，示例样本的基本信息如表 5-1 所示。

表 5-1 示例样本基本信息

| 舞弊类型 | 股票代码 | 证券名称 | 所处行业 | 输出概率 |
| --- | --- | --- | --- | --- |
| 正常样本 | 000006 | 深振业 A | 房地产 | 0.0429 |
|  | 000009 | 中国宝安 | 综合 | 0.0013 |
|  | 000683 | 远兴能源 | 化学原料及化学制品制造业 | 0.0309 |
| P2503 | 000027 | 深圳能源 | 电力、热力生产和供应业 | 0.9910 |
| P2501、P2503 | 002584 | 西陇科学 | 化学原料及化学制品制造业 | 0.9644 |
|  | 600251 | 冠农股份 | 农副食品加工业 | 0.9724 |
| P2506 | 000831 | 中国稀土 | 有色金属 | 0.9420 |

提取后的结果如图 5-1、5-2 所示，图 5-1 为正常样本的梯度类激活映射图，图 5-2 为舞弊样本的梯度类激活映射图。图片经过比例缩放及拉伸之后，覆盖在原始输入灰度图像之上，从左至右依次为财务指标、ESG 指标、内部控制指标。为了更直观地展示不同指标对于预测的贡献程度，本文在一级指标之间插入红色像素列、二级指标之间插入白色像素列对图像进行分割，亮度越亮的区域对最终的预测结果越重要。

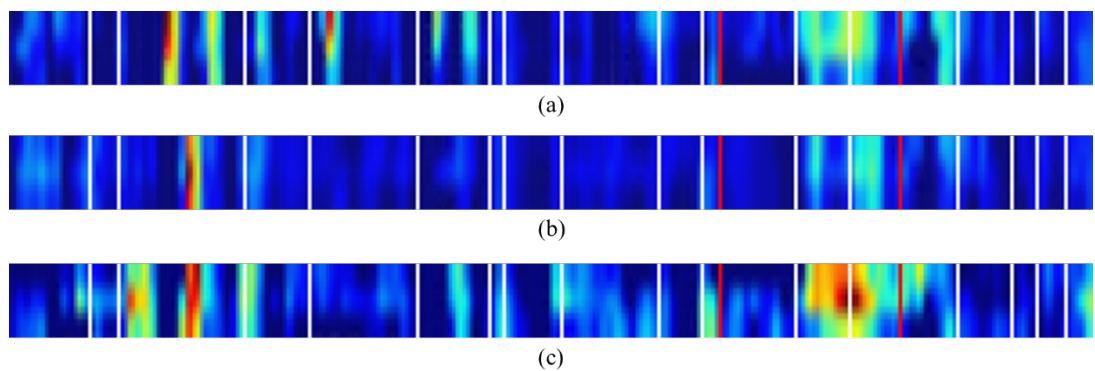

**图 5-1 正常样本最后一个卷积层的梯度类激活映射图**
**(a)深振业 A; (b)中国宝安; (c)远兴能源**

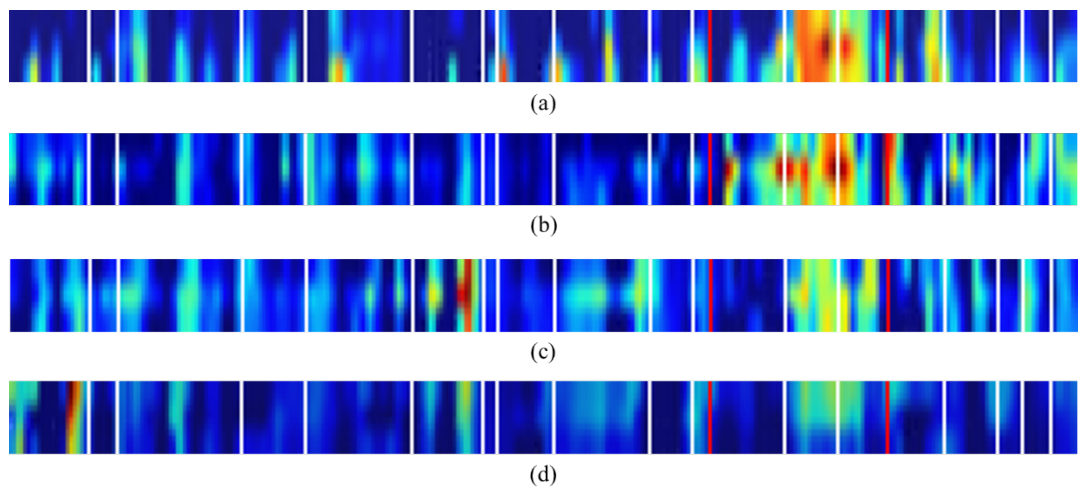

**图 5-2 舞弊样本最后一个卷积层的梯度类激活映射图**
**(a)深圳能源; (b)西陇科学; (c)冠农股份; (d)中国稀土**

观察所有图像，可以清晰地看出，虽然本文的目标变量是企业是否财务舞弊，但非财务指标对于财务舞弊的识别同样有着积极的作用。我们还可以看出神经网络在判别非舞弊企业与舞弊企业时的逻辑共性和差异：

1. 共性

    a) 偿债能力、比率结构、治理结构、内部控制环境这四个二级指标在所有列举出的梯度类激活映射图中都有较为突出的亮度，不论对于非舞弊企业与舞弊企业，这些二级指标都是评价一个公司经营稳定性和治理能力的重要指标。

    b) 环境指标在财务舞弊识别上的作用并不对所有企业显著，而是对某些特定行业的企业作用较大，如上图所列样本的梯度类激活映射图中，只有属于化学原料及化学制品制造业的远兴能源、西陇科学，这两所企业代表环境指标的区域较亮。

2. 差异

    a) 特征维度：非舞弊企业尽管体量不同，但特征模式基本相似，起到主要作用的二级指标为比率结构、经营能力、社会责任、治理结构；而舞弊企业根据舞弊类型差异较大。

    b) 时间维度：非舞弊样本，特征作用模式多为"贯穿式"，即重要特征的各年份数据对于舞弊识别都有作用，在类梯度激活图上表现为亮

色区域多为纵向条带状分布。而对于舞弊样本，特征作用模式多为"团簇式"，即重要特征往往聚集出现在某一个小范围区域，而且作用的时间跨度并不是贯穿取样区间的所有年份，在类梯度激活图上表现为亮色区域多为块状分布。

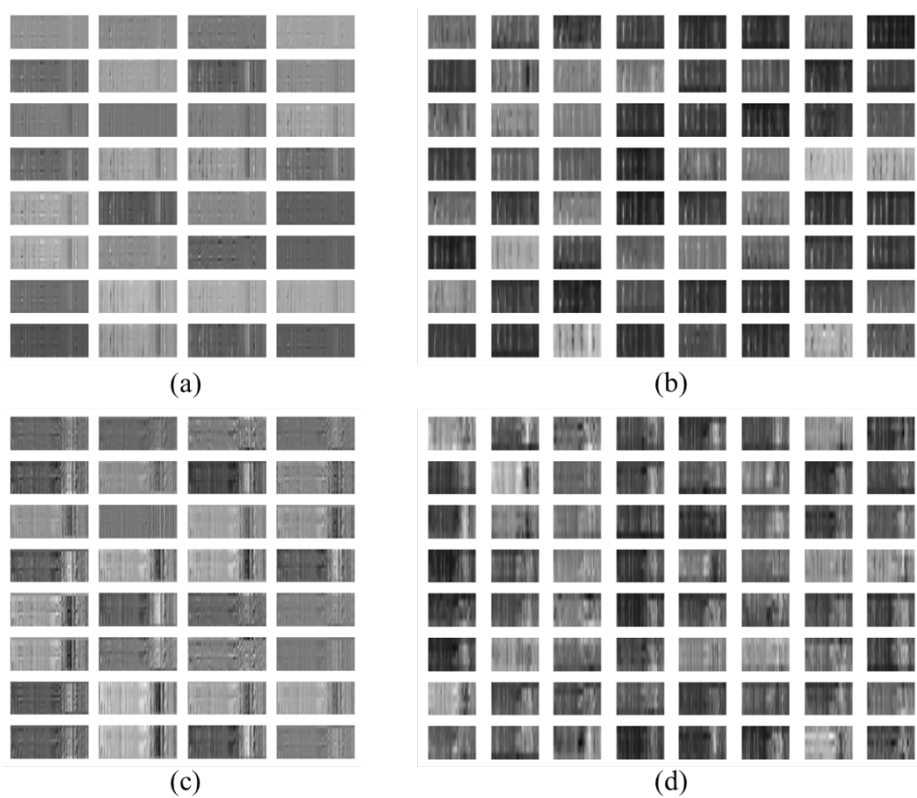

图 5-3 卷积层表征可视化图
(a)深振业 A 第一个卷积层; (b)深振业 A 第四个卷积层;
(c)西陇科学第一个卷积层; (d)西陇科学第四个卷积层

正常样本与舞弊样本的上述差异，还可以通过可视化不同深度卷积层学习到的表征图来佐证。如图 5-3(a)为正常样本第一个卷积层的表征图，5-3(b)为正常样本第四个卷积层的表征图，可以看出正常样本在网络的浅层学习到的特征多为二级指标，在图中的表现为数条紧挨着的纵向粗竖线；在网络的深层，不同卷积核学习到的特征则细化为三级指标，在图中的表现为一条条纵向细竖线。如图 5-3(c)为舞弊样本第一个卷积层的表征图，5-3(d)为舞弊样本第四个卷积层的表征图，可以看出舞弊样本在网络的浅层就已经锁定了该企业的舞弊路径，在网络的深层，"团簇式"特征表现得更为明显，并且不同卷积核分别学习到了不同年份的

特征，在图中的表现为同一条纵向粗竖线的不同亮色分段。

舞弊样本表现出来的上述特征与证监会出具的舞弊判罚条例对应，详见本文 5.2 节的案例分析。

## 5.2 单个样本点案例分析

本文选择冠农股份（股票代码 600251）作为典型舞弊案例，结合中国证监会、上海证券交易所的行政处罚决定与公告进行具体分析。

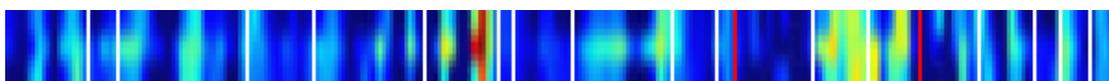

**图 5-4 冠农股份最后一个卷积层的梯度类激活映射图**

冠农股份 2022 年的舞弊类型为 P2501 虚构利润、P2503 虚假记载，借助梯度类激活映射图，卷积神经网络在输出某个样本舞弊概率及预测舞弊情况的同时，还可以输出导致该样本发生舞弊行为的可能原因，为审计人员提供审计路径的指导。由图 5-4 所示，对于冠农股份 2022 年舞弊行为的判定，贡献最大的二级指标为现金流分析、社会责任、治理结构，此外，较为特殊的是，该公司每股指标也十分突出。

中国证监会和上海证券交易所对于冠农股份的违规行为主要认定如下：《2021 年年度报告》存在虚假记载，虚增营业收入 3.53 亿元，营业成本 3.53 亿元；《2022 年半年度报告》存在虚假记载，虚增营业收入 17.28 亿元，虚增营业成本 17.28 亿元。

经查，该公司实现上述舞弊行为的手段主要为收入确认方式不当、收入确认跨期。具体的方式包括：第一，时任冠农股份子公司天沣物产执行董事、法定代表人王新时，被认定存在负责组织天沣物产相关人员修改原始单据的行为；第二，该公司内部控制存在重大缺陷，该公司针对子公司新疆冠农天沣物产有限责任公司 2021 年开展的皮棉贸易业务，修改了部分皮棉贸易销售合同、货权转移证明及费用说明中相关内容，将修改后的文件替换原始文件，并据此确认相关收入，导致你公司披露的 2021 年年度报告存在虚增收入情形（新疆监管局《行政处罚

决定书》【2023】4 号、上海证券交易所《纪律处分决定书》【2023】184 号）。二级指标现金流分析、治理结构、三级指标内部控制在最后一层卷积层梯度较大，说明神经网络能够识别到直接导致财务舞弊的信息。

此外，对于处罚决定书中未提到，但梯度类激活映射图中亮度较高的两个二级指标每股指标、社会责任，本文在溯源后也发现了其对于财务舞弊识别的潜在作用。由图 5-4 所示，每股指标的高亮区域集中在时间维度的中间偏后部分，这是因为在 2018 年，冠农股份被披露公司与控股股东人事管理不完全独立、股份转换后控制的部分子公司与上市公司存在同业竞争情形（新证监函【2019】306 号）。

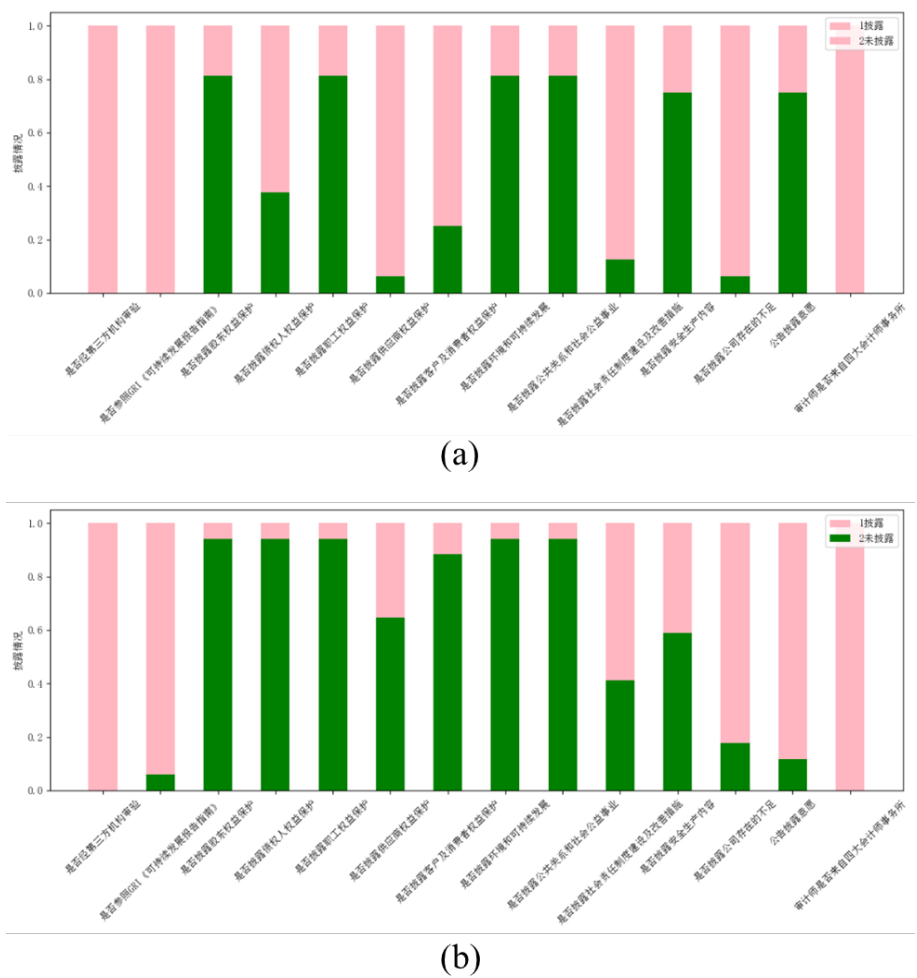

**图 5-5 社会责任二级指标原始数据分布图**
**(a)正常样本：深振业 A; (b)舞弊样本：冠农股份**

本文检视了冠农股份社会责任二级指标下的类别变量，并与正常样本深振业

A 对比。如图 5-5 所示，冠农股份社会责任二级指标下的类别变量的取值中，取值为 2（未披露）的比例要远高于深振业 A。因此本文认为，社会责任披露不积极的企业容易产生舞弊行为，需要重点关注。

最后值得注意的是，冠农股份的舞弊行为发生在 2022 年，中国证监会和上海证券交易所对于冠农股份舞弊手段及具体方式的查证公告均为 2023 年出具并公布的，在舞弊行为发生之后才进行溯源查证。而本文构建的卷积神经网络，利用财务舞弊案件大多数都为多年连续、有组织、有预谋的特征，使用 2010~2021 年的数据，就可以准确预测下一年发生舞弊的可能性，并且给出大致的舞弊路径指示，具有实际应用价值。

# 6 总结与展望

## 6.1 总结

基于财务舞弊危害大、传统方法识别难、机器学习方法可解释性差三点考虑，本文提出使用卷积神经网络识别上市公司财务舞弊，并使用专用于卷积神经网络的模型解释方法对样本进行解释。

本文首先根据文献定义了舞弊样本，并基于现有财务舞弊识别框架构建了包含财务指标、ESG 指标、内部控制指标的舞弊识别指标体系。接着对数据按照股票代码分类，并根据特征的数据类型分别进行缺失值处理，并使用孤立随机森林方法剔除了由于证监会认定滞后性而存在的灰色样本，以及对数据进行标准化处理。之后创新性地将面板数据按照对象维度分割成一张张灰度图像，使神经网络充分利用面板数据时间、特征两个维度的信息。本文还将图像展平进行过采样处理，再还原到原始维度投入网络训练。

本文共定义了两类识别模式的卷积神经网络，分别为事后检测与事前预测，两类网络结构相似、参数略有不同。调参过程中发现，在财务舞弊识别任务中，分类阈值对网络表现有着重要的作用，需要进行特别调整。之后本文将 CNN 与逻辑回归模型、LightGBM 模型进行比较，结果显示事前预测 CNN 预测准确率高、时效性强，并未因缺少一年信息而损失过多预测性能，体现了卷积神经网络的有效性及鲁棒性。本文采用梯度类激活映射图和表征可视化技术，对事前预测 CNN 进行了反向、前向相结合的解释，案例分析直观清晰、符合审计实务直觉，并能提供富有洞察力的审计路径提示。

本文主要有三点贡献：（1）提出了一种可以充分利用面板数据信息、适用于财务舞弊识别领域的深度学习识别方案；（2）明确了分类阈值的确定在财务舞弊识别等高风险领域应用的重要性；（3）采用可视化技术对"黑盒"模型的决策逻辑做出了有说服力的解释。

## 6.2 展望

尽管本文在事前预测财务舞弊事件，以及深度学习可解释性方面取得了一定成果，但本文尚存一定局限性：（1）本文构建的卷积神经网络为二分类网络，仅能预测某个企业在下一年是否舞弊，对于预测为舞弊的企业，无法预测具体的舞弊类型，日后可以考虑使用多分类网络；（2）本文的可解释性研究主要是生成最后一个卷积层的梯度类激活映射图，并将其拉伸放大后覆盖到原始灰度图像上，虽然呈现效果较直观，但拉伸放大的操作会带来一定的模糊性，无法提供像素级别的解释力度，即无法说明某个三级指标在某一特定年份是否可疑，日后可以考虑使用反事实解释方法提供某个具体特征层面的解释；（3）本文的解释只停留在可视化角度，无法提供人类可以直接理解的语义解释，日后的研究中可以探索改变网络结构，融合部分具有自然语言处理的网络层，或引入注意力机制，在输出预测概率的同时输出有指导意义的自然语言。

# 参考文献